\documentclass[10pt,twocolumn,letterpaper]{fairmeta}

\usepackage{geometry}

\newif\ificml
\icmlfalse

\usepackage{xspace}
\RequirePackage{amsmath}
\RequirePackage{amssymb}
\makeatletter
\DeclareRobustCommand\onedot{\futurelet\@let@token\@onedot}
\def\@onedot{\ifx\@let@token.\else.\null\fi\xspace}

\usepackage{balance}

\usepackage{algorithm}
\usepackage{algorithmic}
\usepackage{listings}

\def\eg{\emph{e.g}\onedot} 
\def\ie{\emph{i.e}\onedot}

\def\wrt{w.r.t\onedot} 
\def\iid{i.i.d\onedot} 

\makeatother

\newcommand{\matthijs}[1]{{}} %
\newcommand{\pem}[1]{{}} %
\newcommand{\gsz}[1]{{}} %
\newcommand{\maria}[1]{{}} %
\newcommand{\francisco}[1]{{}} %
\newcommand{\naila}[1]{{}} %
\newcommand{\erv}[1]{{}} %

\def\mypar#1{\vspace{0.5em}{\noindent\bf #1}\hspace{5pt}}
\newcommand{\mysel}[1]{\textcolor{blue!50}{\emph{#1}}}

\usepackage{amsfonts}

\title{Inference-time sparse attention with asymmetric indexing}

\newcommand{\OURS}{\textsc{Saap}\xspace}

\author{Pierre-Emmanuel Mazar\'e}
\author{Gergely Szilvasy}
\author{Maria Lomeli}
\author{Francisco Massa}
\author{Naila Murray}
\author{Herv{\'e} J{\'e}gou}
\author{Matthijs Douze}

\affiliation{FAIR at Meta}

\abstract{
Self-attention in transformer models is an incremental associative memory that maps key vectors to value vectors.
One way to speed up self-attention is to employ GPU-compatible vector search algorithms based on standard partitioning methods such as k-means. However, such partitioning methods yield poor results in this context because (1) the keys and queries follow different distributions, and (2) the RoPE positional encoding hinders the bucket assignment. 

This paper introduces \OURS{} (Self-Attention with Asymmetric Partitions), which overcomes these problems. It is an asymmetrical indexing technique that employs distinct partitions for keys and queries, thereby approximating self-attention with a data-adaptive sparsity pattern. 
It works on pretrained language models and only requires to train (offline) a small query classifier. 
On a long context Llama 3.1-8b model, with sequences ranging from 100k to 500k tokens, \OURS{} typically reduces by a factor of 20 the fraction of memory that needs to be looked-up, which translates to a time saving of 60\% when compared to  FlashAttention-v2. 

}

\date{\today}
\correspondence{First Author at \email{pem@meta.com}}

\begin{document}

\maketitle

\section{Introduction}
\label{sec:intro}

Current transformer-based large language models (LLMs) ~\citep{ Vaswani2017AttentionIA,brown2020language} implement contextual memory in the form of self-attention: The model attends to previously processed tokens in order to predict the next one.
To avoid recomputing the previous keys and values, a key-value (KV) cache is stored and reused, allowing for faster next-token prediction. \nocite{devlin2019bertpretrainingdeepbidirectional,chowdhery2022palmscalinglanguagemodeling}
This KV cache is memory intensive: on a Llama 3-8B model~\citep{dubey2024llama3herdmodels}, a single token involves a cache of size 128~kiB -- about 10000 times larger than storing the token itself.    
Therefore, this cache becomes increasingly memory-demanding as the context length increases.
This problem is partially addressed by quantization to compress key and value vectors \citep{shi2024keep} but comes at a cost, 
as the generation quality eventually degrades below 4 bits per weight. 
Additionally, long-context self-attention requires matching queries one by one to the entire KV cache, which becomes computationally prohibitive as the context grows.

In this work, we address this challenge by adopting a vector search perspective: we regard self-attention as a setting where a query vector is matched with a database of vectors (the key vectors in our case). The goal is to retrieve a subset of relevant vectors. 
The self-attention softmax can then be computed on this restricted set of $k$ vectors, yielding an approximation that improves in accuracy as $k$ increases.
An efficient implementation of this retrieval with exhaustive search %
can improve the speed only to some extent, 
since all keys still need to be accessed even if only $k$ values are used. 

Instead, we advocate approximating nearest-neighbor search by partitioning the keys into buckets. 
Practically, this enables sparse, data-dependent, self-attention: 
at search time, we leverage only a small subset of buckets for each query, reducing memory access, and therefore time complexity.
Multiple choices exist for the partition: random projections, as in the original locality-sensitive 
hashing methods~\citep{datar2004locality}, or data-dependent partitions, such as k-means. The latter offers better trade-offs in practice in typical approximate neighbor search tasks ~\citep{pauleve2010locality}. 
In this approach for retrieving keys, the slow k-means training is performed offline on a held-out training set of keys.  
After this offline phase, and during the pre-fill phase with a new context (and therefore new keys), performing the assignment to a given bucket of the partition is fast, since it amounts to finding the nearest centroid for each new key. %

However, in this context, approximate k-nearest neighbor (ANN) methods to
improve KV cache efficiency face several limitations and challenges: %

\begin{enumerate}
\setlength\itemsep{0em}
\item 
keys and queries have very different distributions, hence vector search is out-of-distribution in this setting. This hinders the effectiveness of indexing algorithms.  %
\item 
retrieving the top-$k$ keys only is restrictive for queries where useful information is associated to keys beyond the top-$k$.  %
\item 
current brute-force attention implementations, such as FlashAttention-v2~\citep{daoflashattention} are highly tuned for the training and inference hardware. In contrast, many ANN algorithms are not GPU-compatible. %
\end{enumerate}

To address these limitations, we propose a novel approach called {\bf S}parse {\bf A}ttention with {\bf A}symmetric {\bf P}artitions (\OURS), that casts the partitioning of key and query vectors as a classification task: the bucket membership is predicted separately for keys and query vectors. %
In addition to this approach, we make the following contributions:%
\begin{enumerate}
\setlength\itemsep{0em}
\item 
     We quantitatively analyze the out-of-distribution nature of query-key matching (Section~\ref{sec:motivation}) and introduce an asymmetrical assignment strategy for keys and queries (Section~\ref{sec:beyondtopk}); 
\item 
     We mitigate the loss of information that occurs when the attention is limited to top-$k$ keys.
     Instead we leverage the partition structure (Section~\ref{sec:beyondtopk}), which leads to a variable size sparsity pattern, and show that this choice improves the self-attention approximation; 
\item 
     We propose a method for efficiently running the partition-based search on GPUs, based on an optimized batched implementation (Section~\ref{sec:implem}). 
\end{enumerate}

\OURS reduces the resources needed to compute the self-attention of a model without finetuning it, in a hardware-compatible manner. Compared to the state-of-the-art FlashAttention-v2 baseline, it decreases the time taken by self-attention kernels by more than 60\%, without compromising the generation quality, see Section~\ref{sec:experiments}.

\section{Related work}
\label{sec:related}

\begin{table}
    \centering
    \caption{\OURS \emph{v.s.}  comparable KV-indexing methods: fast implementations like flashattention-v2 correspond to the full attention baseline. The other columns correspond to  indexing methods that induce sparse attention. \label{tab:sparsemethods}}
    {\small
    \begin{tabular}{@{\ }l@{\ \ }|@{\ \ }cccc@{\ }}
    \toprule
           & \begin{minipage}{1.2cm}\ \ \ Full  \\
           attention\end{minipage} %
           & LSH %
           & \begin{minipage}{1.2cm}Retrieval \\
           Attention\end{minipage} %
           & \OURS \\
           \midrule
    Frozen LLM         & \checkmark & \checkmark & \checkmark  & \checkmark \\  
    Sparse attention   &            & \checkmark & \checkmark  & \checkmark \\
    Fast index build   & \checkmark & \checkmark &             & \checkmark \\
    GPU compliant      & \checkmark & \checkmark &             & \checkmark \\
    Data adaptive      & \checkmark &            &  \checkmark & \checkmark \\
    Approximation      & exact      & medium     & good        & good  \\ 
    \bottomrule
    \end{tabular}}
\end{table}

This section reviews related work on accelerating attention for autoregressive transformers. Table~\ref{tab:sparsemethods} compares our method with a selection of the most relevant approaches along with their characteristics.

\mypar{KV cache compression.}
Multiple techniques address the large size of the KV cache for long sequences. 
Some approaches rely on vector compression and pruning, similar to neural network pruning \citep{lecun1989optimal}  and compression~\citep{han2015deep}.
The most straightforward way is to adopt low-precision computation variants of scalar quantization for the KV cache.
This is effective until the number of bits per weight becomes detrimental (around 4 bits) and the compression severely degrades the accuracy~\citep{li2024evaluating}, even when the compression is accounted for during training~\citep{adepu2024framequant} with quantization-aware training~\citep{hubara2018quantized}. 
The keys and values can also be compressed by dimensionality reduction, \eg with Principal Component Analysis~\citep{kang2024gear}.

Another approach to compress the KV cache is to prune the least-used keys and vectors by measuring their utilization on-the-fly~\citep{ge2023model}, maintaining a cache of the most used vectors~\citep{liu2024scissorhands}, ignoring the oldest keys~\citep{xiao2024efficient}, or using a heuristic such as H2O~\citep{zhang2023h2o}.
The GEAR method~\citep{kang2024gear} combines pruning, dimensionality reduction and vector compression.
For more context on KV cache compression and compute optimization, we refer the reader to the recent overview by \citet{shi2024keep}.
Our \OURS approach is complementary to compression and static pruning. 
Indeed, partition-based vector search techniques are often combined with compression, for instance, the canonical IVFADC method~\citep{jegou2010product}.

\mypar{Vector search data structures.}
Several families of vector indexing structures have been applied to KV cache indexing.  
In the vector search literature, the ``keys'' are called ``database'' or ``reference'' vectors. 
Tree-based indexes are inspired by one- and low-dimensional search such as the seminal KD-tree~\citep{bentley1975multidimensional}, 
but are not effective in high dimensions. 
 Locality Sensitive Hashing (LSH) is based on hash tables where buckets store the database vectors~\citep{datar2004locality}. 
The advantage of LSH is its fast indexing time, as it amounts to computing a limited number of dot products for each key. 
This advantage has led to its use in several recent works on KV cache indexing~\citep{zandieh2023kdeformer,zhuoming2024magicpig}. 
In Appendix~\ref{app:lshvariants}, we discuss LSH-based methods and show that it performs poorly for KV cache acceleration: vanilla LSH is not data-adaptive, so the number of independent hash tables has to be increased significantly to achieve a decent retrieval accuracy. This consumes a lot of memory and overall does not provide a significant gain with pretrained models.  

Graph-based indices store database vectors as nodes and searching hops from node to node~\citep{dong2011efficient,malkov2018efficient,fu2017fast,jayaram2019diskann}. \nocite{chen2024roargraph,ootomo2024cagra}
Graph-based search is efficient because routing decisions are built into the graph during its construction. 
Therefore, RetrievalAttention~\citep{liu2024retrievalattention} applies graph-based index RoarGraph~\citep{chen2024roargraph} to the KV cache. 
The downsides are that the index building is slow and the graph is bulky to store, as each node is linked to dozens of neighbors. 
As a result, graph indexing of a KV cache is attractive only if the same KV cache is reused, \eg because multiple prompts use the same context. 

\newcommand{\nlist}{C}
\newcommand{\nprobe}{\ell}

\mypar{Vector search based on partitions.}
With this approach the reference dataset is partitioned into $\nlist$ buckets.
Partitions are defined by a function that assigns each vector to one bucket. 
The classical variant employs an inverted file (IVF) \citep{witten1999managing}. 
When the partitioning is based on k-means, as advocated by \citet{pauleve2010locality}. The usage is as follows: 
(1) \emph{At training time}, k-means clustering of the $N$ database vectors provides a set of $\nlist$ centroids. 
(2) \emph{At indexing time}, the assignment of a new vector amounts to finding its nearest centroid. 
Ideally, the IVF structure stores the vectors associated to one bucket in contiguous memory.   %
(3) \emph{At search time}, the query vector is likewise assigned to the nearest centroid and all the database vectors found in the bucket are compared with the query vector to find the top-$k$ nearest neighbors. 
A straightforward extension is to visit \emph{several} 
nearest buckets instead of the single nearest one \citep{lv2007multi}. 
The NeuralLSH method~\citep{dong2019learning} predicts the bucket corresponding to the queries and database vectors alike using a neural network. 
Unlike our approach, it does not handle out-of-distribution data. 

Vector search tooling such as Faiss has been used for KV cache lookups in the Unlimiformer method~\citep{bertsch2024unlimiformer}, in the setting of an encoder-decoder LLM where the entire decoder shares a single cache, independently of the layer. 
This factorization is not possible in decoder-only LLM architectures, where each layer and each head need to store the (key, value) pairs independently.

\mypar{Out-of-distribution (OOD) search.}
By default, vector search assumes query and database vectors follow the same distribution. 
OOD search is needed when it is not the case, for example, if the query and database vectors come from different modalities as with text-to-image search~\citep{simhadri2024results} or user-to-item search in recommendation systems~\citep{paterek2007improving}.
The effects of OOD search on partition-based indexes, as identified by \citet{jaiswal2022ood} and \citet{chen2024roargraph}, are: 
(1) the nearest database vectors to the query are spread over many buckets, and 
(2) the query assignment does not visit the correct buckets.
The authors adapt graph-based search to OOD search by taking into account how the graph navigation must be re-routed, based on a training set of query vectors. 

KV cache indexing has challenging OOD characteristics that we discuss in Section~\ref{sec:motivation}, and that our method \OURS addresses in a principled way in  Section~\ref{sec:method}. 
Dedrift~\citep{baranchuk2023dedrift} addresses temporal drift in vector databases for partition-based indexes by re-training mid-way during key ingestion. 
Our \OURS approach mitigates the temporal biases in a simpler way, see Section~\ref{sec:derope}.

\begin{figure*}[t]
    \includegraphics[height=0.3\linewidth]{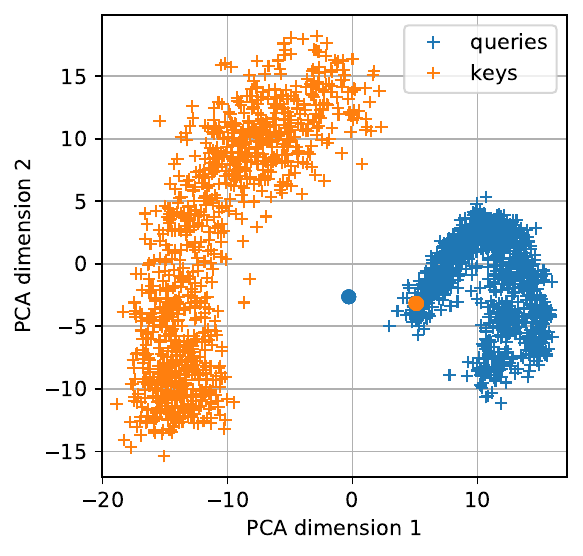}
    \hfill
    \includegraphics[height=0.3\linewidth]{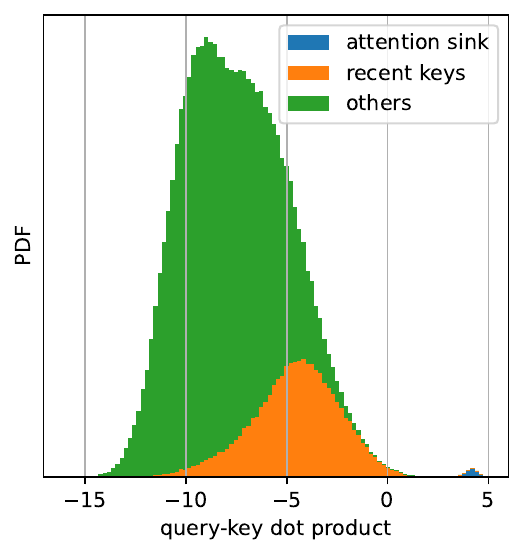}
    \hfill
    \includegraphics[height=0.3\linewidth]{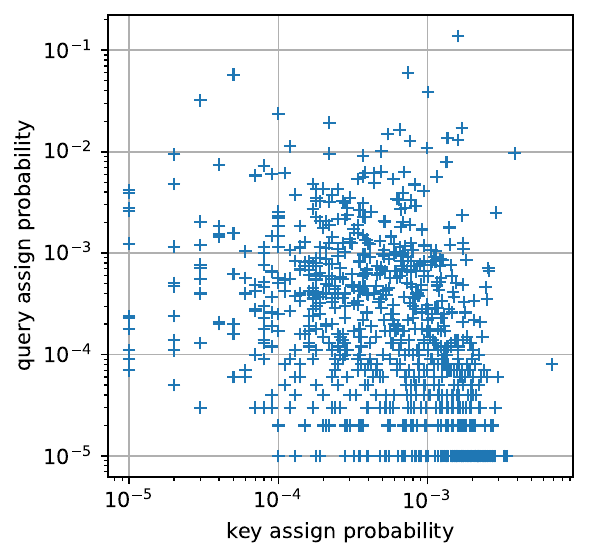}
    \caption{
    Illustration of the OOD statistics between keys and queries in the self-attention. 
    \emph{Left:} first two PCA dimensions of keys and queries (the big dots are the attention sinks for keys and queries). 
    \emph{Middle:} distribution of dot products between random subsets of keys and queries.     
    \emph{Right:} after clustering keys, probability of assignment to each cluster for  keys (x-axis) and queries (y-axis).
    \label{fig:ood}}
\end{figure*}

\section{Preliminary discussion}
\label{sec:motivation}
\label{sec:derope}

In this section, we review a few key properties of the attention operator in LLMs from a vector search perspective. 
We manipulate query vectors (token query embeddings) and key vectors (token key embeddings), ignoring the value vectors for now. 
In a transformer model, queries are matched to keys from a given attention head.
Therefore, we train and build indexes separately for each head.

\mypar{Attention as a vector search application}
Considering the self-attention computation,
the main differences with respect to classical approximate nearest-neighbor (ANN) settings are:
(i) there is not a hard cut-off for relevant results, but a soft weighting of results using softmax; 
(ii) it is a maximum inner product search problem instead of the more common Euclidean or cosine-similarity search; and
(iii) the keys and queries are produced by different computations, thus, they reside in different embedding spaces.
To accelerate the computation, we approximate exhaustive attention using a hard cut-off on the maximum number of retrieved keys.

\mypar{Out-of-distribution setting and partitioning of keys}
Classical vector search methods are designed to find vectors that are sampled \iid from the same distribution. 
We examine two types of OOD behavior: (i) between keys and queries, and (ii) between training and testing. 
Prior work has quantified the difference between key and query distributions by comparing histograms of query-to-key distances v.s. key-to-key distances~\citep{jaiswal2022ood,chen2024roargraph}. 
\OURS is based on a partition of the keys. 
For a query vector, the $\nprobe$ most promising buckets are visited and the dot product between the query and all keys from the visited buckets is computed. 
One important aspect that plays into OOD behavior for \OURS, is the degree to which the buckets of the $N$ key vectors are balanced in terms of membership. 
If the buckets are all of the same size and equally likely to be visited, the number of dot products to perform is $\nprobe N / \nlist$. 
The search time is proportional to this.

\mypar{Key-query OOD}
Keys and queries are computed using distinct linear layers on top of shared vectors.
The  distribution mismatch between keys and queries is observed with basic statistics of a random sample of vectors (here head 28 of layer 17 of a Llama 3 model). %
\autoref{fig:ood} (\emph{left}) shows the first principal components of keys and queries: their embedding spaces are clearly distinct. 
As the two point clouds are on opposite sides of the origin, the $\langle k,q \rangle$ dot products are almost always negative (\autoref{fig:ood}, \emph{middle}).
The dot products with the first key (\ie the ``attention sink''~\citep{xiao2024efficient}) and the most recent keys, are statistically higher than the other keys. 

From now on, we apply a spherical k-means clustering with $\nlist\,=\,1024$ centroids to the keys~\citep{douze2024faiss}. 
Figure~\ref{fig:ood} (\emph{right}) shows that the probability of assignment to a given centroid is very different for keys and queries.
More specifically, some clusters ``capture'' a significant fraction of the queries (up to 10\%). 
Motivated by this, we re-design an alternative assignment operator in Section~\ref{sec:method}.

\begin{figure}[t]
    \includegraphics[width=\linewidth]{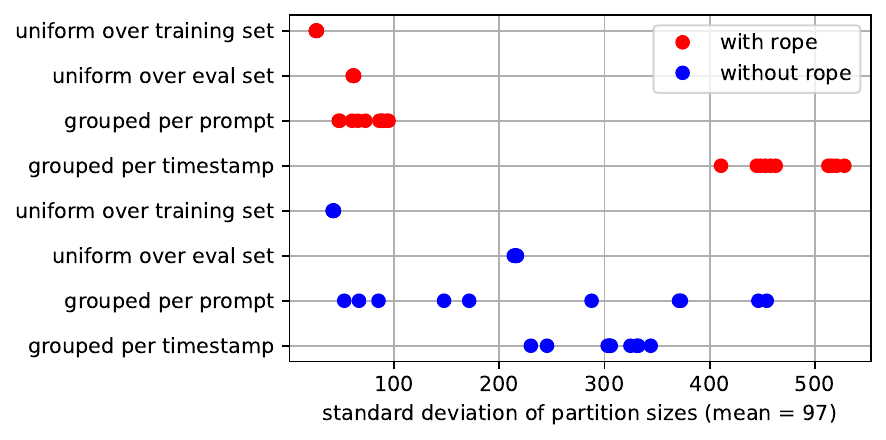} 
        \vspace{0.05em}
    \caption{
Deviation of key bucket sizes when 100k keys are assigned to a trained k-means.
The 100k keys are sampled 10 times, in 4 different ways indicated on the y-axis, and with or without the rope transformation applied.
    \label{fig:otherOOD}}
\end{figure}

\mypar{Prompt and temporal OOD}
\label{sec:otherOOD}
OOD problems occur for all ML models that assume \iid training and test data. 
Here we examine the divergence of the distribution at training time compared to at test time when a prompt is used, focusing entirely on the key vector distribution.
We measure the variance of the assignment of 100k vectors sampled in different ways, and repeat the experiment 10 times: higher variance indicates stronger OOD behavior.
Figure~\ref{fig:otherOOD} shows that the minimal variance occurs when vectors are sampled uniformly from the training set. 
When we measure it on 100k random uniform vectors from data that is disjoint from the training set, the variance increases.
We sampled 100k vectors from each of 10 different prompts and found that the variance changes significantly depending on the prompt: this is prompt OOD. %
We then sample 100k vectors across prompts, but from 10 narrow timestamp ranges: here the deviation with respect to the training distribution is the highest, showing strong temporal bias.

\mypar{Removing temporal bias.}
The rope temporal transformation~\citep{su2023roformerenhancedtransformerrotary} is likely a major contributor to their temporal bias when applied to all keys and queries. 
Therefore, we repeated our analysis on keys without the rope transformation and found that indeed the time bias is reduced (blue dots in Figure~\ref{fig:otherOOD}). 
We then applied the same de-rope operation to queries where the rope component has been removed, yielding the last plot in Figure~\ref{fig:otherOOD}.
Given the positive effect of de-roping, we use it for all models trained for \OURS.

\mypar{Per-head attention span}
\label{sec:densepart}
Figure~\ref{fig:ood} (\emph{middle}) shows that the first key of any sequence, the so-called \emph{attention sink}~\citep{xiao2024efficient}, is an outlier in the data distribution and always has a large dot product with the query vectors. 
Similarly, the most recent keys have higher dot products with the queries. 
Therefore, it is typical in attention approximations to retrieve these keys exactly. 
Approximate and exact retrieval results are then combined to form a final attention result. 
In our case, we use the ``1+2047'' setting, which means that the first token and the most recent 2047 tokens are attended exhaustively. 
This context is long enough to have good performance on many short-context tasks.

\begin{figure}[t]
    \centering
    \includegraphics[width=0.75\linewidth]{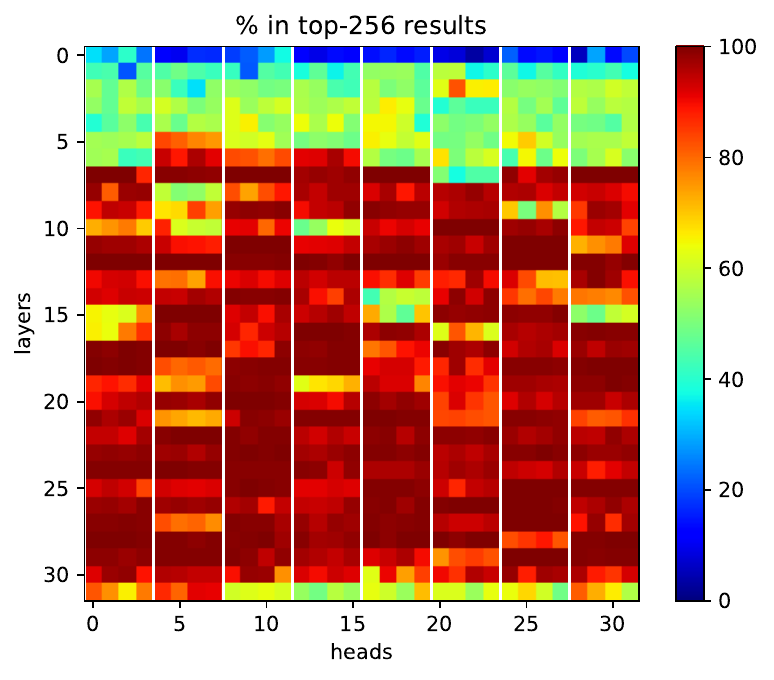}
    \vspace{-1em}
    \caption{Fraction of attention contributed by the top 256 tokens (out of $10^5$), per head, averaged over 10 prompts. 
    \label{fig:attention_fraction}}
\end{figure}

Figure~\ref{fig:attention_fraction} shows the fraction of attention that comes from the top 256 keys (out of $10^5$) for each query (excluding the attention sink and recent tokens). 
For most heads, the top-256 cover most of the attention weight. 
Layer 0 is an exception, all its heads are more uniformly spread over keys. 
Therefore, in the following, we perform full attention for the layer 0, similarly to the work by \citet{tang2024quest,xiao2024efficient}. 

\section{Training and searching partitions}
\label{sec:method}
\newcommand{\nq}{n_{\mathrm{q}}}
\newcommand{\nk}{n_{\mathrm{k}}}
In this section, we present more details on how our \OURS method assigns keys and queries to buckets. %
Note that a key will be assigned to a single partition bucket (unlike LSH). 
 On the other hand, queries are assigned to the $\nprobe$ most promising buckets, in a multi-probe fashion.
The assignment function aims at assigning key vectors  to a relevant bucket and finding the top-$\nprobe$ best buckets for the queries.
A common solution is to partition the data with LSH or clustering. 
Clustering is more effective than LSH variants (see Appendix~\ref{app:lshvariants}), which are not data-adaptive. 
Thus, we use standard k-means~\citep{jegou2010product} for clustering $K\in \mathbb{R}^{\nk \times d}$ keys into $\nlist$ centroids. 
K-means has the advantage that it produces relatively balanced clusters. 

\mypar{Asymmetric assignment.} 
The first adaptation we perform is to address the key-query OOD setting (see Section~\ref{sec:motivation}). 
For this, we separate the assignment function for keys and queries: for queries we train a classifier $f_\mathrm{q}$ that  outputs a probability distribution over $[0,1]^C$. 
At search time, for query $q$, we compute the attention only for the keys in the clusters given by $\mathrm{argmax}_{\nprobe} f_\mathrm{q}(q)$.

\mypar{Assigning with(out) RoPE. }
The second issue with KV cache indexing is the temporal OOD behavior, meaning that long-range lookups that are relatively independent of position are hard to match.  
A large part of this is due to the RoPE transformation of vectors. 
Therefore, we apply the k-means clustering on vectors \emph{before} transforming them with RoPE. %
The keys and queries from which the attention is computed are still transformed with RoPE.

\mypar{Query assignment function}
\label{sec:qpart}
Here we assume that k-means is already trained.
We use a training set of queries $Q\in \mathbb{R}^{\nk \times d}$ and keys $K\in \mathbb{R}^{\nk \times d}$. 
The hard assignment is represented as a binary matrix $H_\mathrm{k} = \{0, 1\}^{\nk \times C}$,  where entry $(i, j)$ is set to 1 if key $i$ was assigned to bucket $j$.
Then, the entry $(i, j)$  of the matrix $A \times H_\mathrm{k}$ represents the fraction of attention weight for query $i$ contained in bucket $j$.
Therefore, the output distribution of $f_\mathrm{q}$ should be as similar as possible to row $i$ of $A \times H_\mathrm{k}$. 
This yields the following loss: 
\begin{equation}
    \mathcal{L}_\mathrm{q} = 
    \mathrm{KLDiv}\left(f_\mathrm{q}(Q), A H_\mathrm{k}\right), 
    \label{eq:kldiv}
\end{equation}
where $\mathrm{KLDiv}$ is the Kullback-Leibler divergence between distributions, and $\mathrm{KLDiv}$ and $f_\mathrm{q}$ are applied row-wise on their matrix arguments. 
Each training batch comes from a separate prompt. 
We sample $\nq$ queries $Q$ and $\nk$ keys $K$ uniformly from the prompt. 
Since short-term queries are taken into account by the ``1+2047'' dense part, we force the training to focus on long range queries by sampling only (key, query) pairs that are more than 2047 steps apart. 
See Appendix~\ref{app:trainingdetails} for details about the training parameters.

\mypar{Beyond top-k. }
\label{sec:beyondtopk}
The IVF approach retrieves only the top-$k$ keys that have the largest dot product with the query. 
However, collecting the top-$k$ results in vector search is slow~\citep{johnson2019billion} on GPUs. 
Besides, retrieving the top $k$ does require computing the dot products \wrt \emph{all} keys within a matching bucket anyways. 
Therefore, we aggregate \emph{all} the corresponding values into the attention result on-the-fly.

\section{Implementation}
\label{sec:implem}

\mypar{Batched queries.}
Llama 3 8B models use group query attention~\citep{ainslie2023gqatraininggeneralizedmultiquery}, where four heads from one layer share the same KV cache. 
At search time, for each token, we need to perform 4 searches for queries $\{q_1, q_2, q_3, q_4\}$. 
For efficiency, we perform the 4 searches jointly by combining the outputs of the query assignment model as $\sum_{i=1}^4 f_\mathrm{q}(q_i)\in\mathbb{R}^{\nlist}$, and pick the top-$\nprobe$ values from this vector. 
This approximation converts a matrix-vector product into a matrix-matrix multiplication (albeit with a small batch size of 4). 
In Section~\ref{sec:ablations} we measure the impact of this approximation.

\mypar{On GPU.}
The \OURS kernel follows an implementation similar to the Triton-based Flash-Decoding~\citep{dao2023flash} included in xFormers~\citep{xFormers2022}. In contrast to Flash-Decoding, which splits the keys and values uniformly across different CUDA blocks, in our implementation, each CUDA block computes the partial attention for a single cluster assignment, with the dense local attention (Section~\ref{sec:densepart}) occurring in its own CUDA block.
The custom kernel is necessary in order to limit expensive synchronizations between CPU and GPUs due to the variable sizes of the visited clusters. 
It could be further optimized in a way to similar to  \citet{shah2024flashattention3}.

The issue with this approach is that the runtime for one layer depends on the largest cluster that is accessed by any of the heads in the layer (\ie the maximum over 
$8\times\nprobe$ clusters). 
Therefore, we rely on the key partitioning to produce balanced clusters. 
Note that the KDEFormer and Reformer enforce the clusters to be artificially balanced, this has a significant impact on accuracy (see Appendix~\ref{app:lshvariants}).

\section{Experiments}
\label{sec:experiments}

In this section, we evaluate \OURS on two natural language processing (NLP) benchmarks and report end-to-end LLM performance metrics. We provide a comparison with the most important baselines. Appendix \ref{app:lshvariants} provides an additional comparison with LSH partitioning, while Appendices \ref{app:rulerresults} and \ref{app:infinitebenchresults} report additional results on the Ruler \citep{hsieh2024rulerwhatsrealcontext} and InfiniteBench \citep{zhang2024infinitybench} benchmarks, respectively. 

\begin{figure*}
    \includegraphics[width=0.45\linewidth]{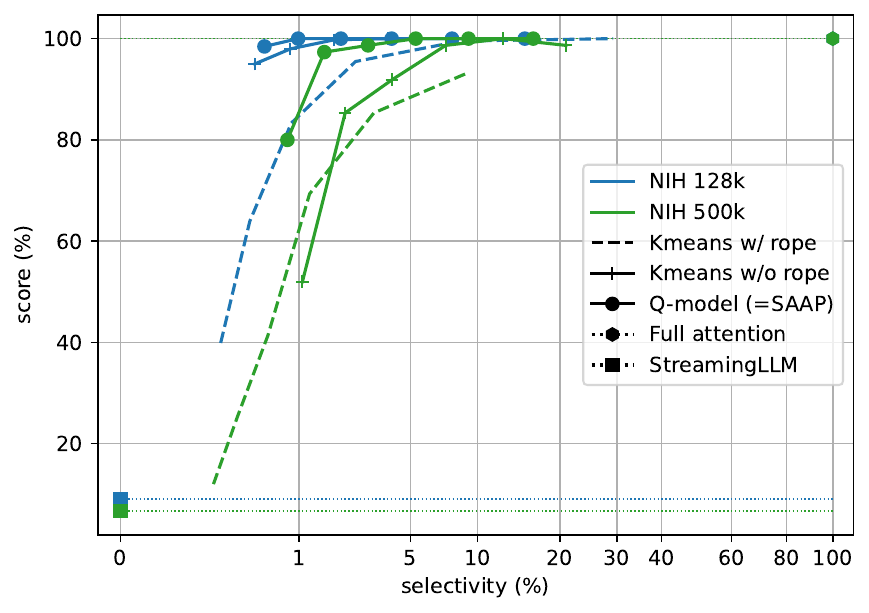}
    \hfill
    \includegraphics[width=0.45\linewidth]{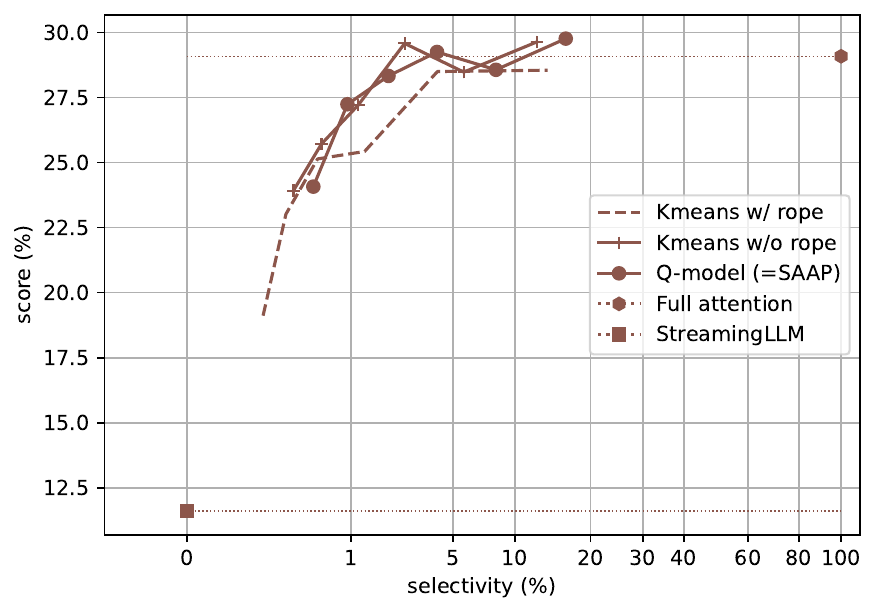}
    \caption{
    Scores of Needle-in-haystack (left) and the En.QA task from InfiniteBench (right) as a function of the selectivity (fraction of KV-cache visited), on a Llama 3 8B with scaled rope to 500k. 
    We compare several ways of assigning keys and queries to buckets.
    We also show the score with no sparse attention (StreamingLLM) and the full attention score. 
    \label{fig:accuracyatbudget}}
\end{figure*}

\subsection{Models, measures, benchmarks and baselines}

We perform experiments using a Llama 3.1 8B, a LLM with 32 layers, 32 query heads and 8 key heads.
This model is trained for context sizes up to 128k tokens. 
We extend this size to 500k tokens using rope scaling~\citep{peng2023yarnefficientcontextwindow}. 
Our method does not require to fine-tune the model weights. 

\OURS first partitions the keys with a pre-trained k-means and trains a Multilayer Perceptron (MLP) in order to assign queries to buckets (offline), denoted by \textit{K-means+Q-model}. The MLP has two layers with an intermediate dimension of 1024, batch normalization and a ReLU nonlinearity.
We use the \OURS assignment models on the PG19 books dataset~\citep{raecompressive2019} for training the MLP.
This data contains natural-language text, but may be out-of-distribution for math or coding tasks.
We sample key and query vectors across the prompt for training, and keep only long-range queries: we discard all queries whose distance to the top-1 matching key is less than 1024 tokens.
We use frozen key and query embeddings for training. See Appendix~\ref{app:trainingdetails} for more training details. 

\mypar{Performance metrics: selectivity and generation timings. }
The main goal of \OURS is to speed up the attention computation and, consequently, decoding.
Therefore, we measure and report resource consumption as a trade-off between accuracy and computing speed. 
We measure the computing speed in two ways: the first is \emph{selectivity}~\citep{zhuoming2024magicpig,pauleve2010locality}, or the rate of keys that are involved in the computation, averaged over all heads; the second is a direct \emph{timing} of the end-to-end generation.

\mypar{Benchmarks.}
We evaluate the end-to-end accuracy on long-context tasks that require different types of lookups~\citep{li2024evaluating}, more precisely: 
 \begin{itemize}
 \setlength\itemsep{0em}
\item 
\textit{Needle-in-a-Haystack}~\citep{NIHGithubURL} consists of a ``hard'' text matching task where a secret, short text (the needle) is inserted at an arbitrary position in an arbitrarily long context made of book texts (in-domain for the training data). 
The prompt attempts to retrieve the text.
We average the %
scores over prompt lengths between 10k and 128k,  reporting separately for lengths between 128k and 500k, and for various insertion points for the needle.  %
\item
\textit{InfiniteBench benchmark}~\citep{zhang2024infinitybench} is a series of tasks designed to probe LLM reasoning capabilities beyond a context size of 100k. 
The tasks relate to question answering, coding and math questions, as well as retrieval tasks for numbers, passkeys and KV. %
\end{itemize}

\mypar{Baselines.} 
We compare \OURS to the following methods:  \\[-1.8em]

\begin{itemize}
 \setlength\itemsep{0em}
\item 
\textit{Full attention} is the default brute-force attention computation. 
We use FlashAttention-v2~\citep{daoflashattention}, a strong open-source reference implementation for long-context attention.  
\item 
\textit{K-means} refers to a partition-based approximate attention using k-means partitioning.
We report results with k-means directly applied to the full keys and queries, as well as applying it to keys and queries without the rope transformation.
This is similar to Faiss' IVFFlat index \citep{douze2024faiss}, which has been used as a baseline by \cite{liu2024retrievalattention}, but in our case it is not restricted to the top-k nearest keys (see Section~\ref{sec:beyondtopk}: beyond top-k). 
\item 
\textit{StreamingLLM}~\citep{xiao2024efficient} corresponds to setting $\nprobe=0$ in IVFFlat, using only the dense part of the attention computation. 
\end{itemize}

\subsection{Comparison with baselines}
\label{sec:baselines}

\begin{table*}
    \caption{
    Scores with a Llama 3.1-8B model using rope extrapolation to increase the context length of 500k and different partitioning schemes. We report in \textbf{bold} the best results among approximate attention schemes. All methods involving k-means assignment use $\nprobe$\,=\,$32$ for multi-probing.  %
    \label{tab:scores}
    }
\noindent \centering  \scalebox{0.88}{
\begin{tabular}{@{\ }l|r@{\ \ \ }r|r@{\ \ \ }r@{\ \ \ }r@{\ \ \ }r@{\ \ \ }r@{\ \ \ }r@{\ \ \ }r@{\ }}
    \toprule
    Method & \multicolumn{2}{c|}{NIH} & \multicolumn{7}{c}{InfiniteBench~\citep{zhang2024infinitybench}} \\
& 128k & 500k & Retr.N& Retr.P&Retr.KV& Math.F & Code.D & En.QA & En.MC\\
    \midrule
\textcolor{gray}{Full attention}  & \textcolor{gray}{100.00} & 	\textcolor{gray}{100.00} &\textcolor{gray}{100.00}  & \textcolor{gray}{100.00} & \textcolor{gray}{21.40}	& 	\textcolor{gray}{30.29} &\textcolor{gray}{31.98}&\textcolor{gray}{29.08}&\textcolor{gray}{48.03}\\
K-means (roped inputs) 
              & 95.48 & 69.33	 &98.98& 100.00 &4.40& 26.00&31.47&25.42& 48.03 \\
K-means (no rope)
              & 100.00 & 98.67	 & 99.66 & 100.00 & 4.40&\bf{34.29}&31.47&29.58&50.66  \\
StreamingLLM (1+2047) & 9.05 & 6.67 &1.69  & 1.69	 & 0.8	 &20.29 &\bf{31.97}&11.60&46.72 \\
\midrule
\OURS, K-means + Q
& \bf{100.00} & 	\bf{100.00} & \bf{100.00} &\bf{100.00}& \bf{7.40}& 32.57&30.20&\bf{29.68}& \bf{50.21}\\
\hfill \mysel{selectivity $\rightarrow$}
& \mysel{4.0} & \mysel{5.3}	 & \mysel{15.0} & \mysel{13.8} & \mysel{21.9} & \mysel{47.8} & \mysel{4.8} & \mysel{3.7} & \mysel{3.7} \\
    \hline
  \end{tabular}}
\end{table*}
\begin{table*}[t]
    \centering
    \caption{
    Comparison of \OURS ($\nprobe$\,=\,$32$) with state of the art methods. 
    The base model is a Llama 3 8B finetuned by Gradient AI for context length of 262k.
    *: \emph{rows reported from~\citet{liu2024retrievalattention}}.
    }
    \smallskip
\scalebox{0.88}{
  \begin{tabular}{@{\ \ }l|r@{\quad}r@{\quad}r@{\quad}r@{\quad}r@{\quad}r@{\quad}r@{\ \ }}
    \toprule
    &  \multicolumn{7}{c}{InfiniteBench~\citep{zhang2024infinitybench}} \\
    Method &  Retr.N& Retr.P&Retr.KV& Math.F & Code.D & En.QA & En.MC\\
    \midrule
    \textcolor{gray}{Full attention*} & \textcolor{gray}{100.0} &\textcolor{gray}{100.0}&\textcolor{gray}{17.5}& \textcolor{gray}{19.0} &\textcolor{gray}{39.5} & \textcolor{gray}{9.1} & \textcolor{gray}{68.0}\\
    StreamingLLM~\citep{xiao2024efficient} * &5.0 &5.0 &1.0&18.5 &39.5 &5.9 &66.5\\
    SnapKV~\citep{li2024snapkv} * &100.0 &100.0&0.5&18.0 &40.0 &11.8 &67.0\\
    InfLLM~\citep{xiao2024infllm} * &100.0 & 100.0&0.5&20.5 &48.0 &7.0 &37.0\\
    RetrievalAttention~\citep{liu2024retrievalattention} * & 100.0 &100.0&9.0/14.0& 19.0 & 40.0 & 7.5 &67.0\\
    \midrule
    \textcolor{gray}{Full attention (reproduced)}   &\textcolor{gray}{100.0}&\textcolor{gray}{100.0}&\textcolor{gray}{16.0}& \textcolor{gray}{41.5} & \textcolor{gray}{24.5}&\textcolor{gray}{10.3}  & \textcolor{gray}{54.5}\\
    \midrule
    \OURS, K-means + Q  
    &100.0& 100.0 & 5.5 & 38.5 &25.5&11.6&56.0 \\
    \hfill \mysel{selectivity $\rightarrow$}  
    & \mysel{25.8} & \mysel{25.9}  & \mysel{19.1} & \mysel{51.0} & \mysel{4.8} & \mysel{4.3} & \mysel{4.3}\\
    \bottomrule
  \end{tabular}
    \label{tab:SOTA}}
\end{table*}

\autoref{fig:accuracyatbudget} shows the trade-off between budget and score (we report detailed numbers in~\autoref{tab:scores}). 
The operating point is controlled by setting the query-time $\nprobe\in \{0, 4, 8, 16, 32, 64, 128\}$. We also show 1) the $\nprobe=0$ setting (\textit{StreamingLLM} ~\citep{xiao2024efficient}), that corresponds to the bottom line; and 2)
the full attention, where the keys are compared exhaustively, as the topline.

Needle-in-a-Haystack scores reach 100\% accuracy from $\nprobe$\,$=$\,$8$ for context sizes up to 128k, but requires $\nprobe$\,$=$\,$32$ beyond that. 
On InfiniteBench, the behavior depends on the tasks. 
 In~\autoref{fig:accuracyatbudget}, En.QA exhibits a clear positive trend, and the topline accuracy is reached with around 5\% selectivity. 
 
For Code.D and En.MC the results for $\nprobe = 0$ and full attention are close, so the scores are noisy. 
For Math.F, the selectivity increases quickly with $\nprobe$, which is indicative of an OOD behavior between training and testing prompts. See also ~\autoref{fig:ibres} in ~\autoref{app:infinitebenchres}.   
This is because for this task, the context consists of long lists of numbers, which is very different from the books on which the assignment models are trained. 
This causes a large imbalance factor.
Interestingly, the \OURS{} score is higher than the full attention score. 
This may be because the sparse attention focuses only on a relevant subset of the prompt instead of averaging information from the entire sequence.

\subsection{Timings}

\textbf{Generation.} We use 8 H100 GPUs 80\,GB (1 KV-cache per GPU) and measure the attention computation speed for one head on a 171k-length context ($\nprobe$\,=\,$32$, yielding 100\% needle-in-a-haystack accuracy). 
We compare against the FlashAttention-v2 implementation~\citep{dao2023flash}. 
\begin{center}
\begin{tabular}{l|rr}
\toprule
& FlashAttention-v2 & \OURS $\nprobe$=32 \\
\midrule
selectivity  & 100\,\% & 4.4\,\% \\
runtime & 50\,$\mu$s & 18\,$\mu$s \\
\bottomrule
\end{tabular}
\end{center}

The FLOPs necessary to compute the sparse attention are about 20$\times$ lower than for the full attention. 
The reduction of runtime is 65\%, because the computation is not as well optimized as FlashAttention and includes overheads. Note that end-to-end timings do not only depend on the budget, but also on how balanced the clusters are and whether the heads all perform the same number of FLOPs. 
This is due to the parallelization over GPUs and over computation cores that we use (see Section~\ref{sec:beyondtopk}).

\textbf{The training} takes around 25~minutes per head on one GPU. 
This is not a limitation since the training is meant to be re-used across contexts and prompts.

\subsection{Comparison with the state of the art}

We compare \OURS with RetrievalAttention~\citep{liu2024retrievalattention}. For this comparison, we employ  a Llama 3 model fine-tuned by Gradient AI to support 262k contexts. 
We set $\nprobe$\,=\,$32$, yielding a selectivity of around 5\% for textual content. 
 
\autoref{tab:SOTA} shows that the baseline numbers are often different, probably because of slightly different experimental settings (\eg prompts). 
However, the gap between \OURS and the full attention baseline is very similar to the RetrievalAttention results. 
Note that in our experimental setting, the dense part of the attention is 1+2047 tokens, while it is 128+512 for RetrievalAttention. 
However, for long-range tasks, this does not matter much as it is still a tiny fraction of the attention size (see the ablations in Section~\ref{sec:ablations}). 
The critical advantage of \OURS is that the slowest stage (training) is performed offline, while RetrievalAttention's RoarGraph index~\citep{chen2024roargraph} has to be built once the context is available, which takes several minutes for a 100k context. 
As mentioned in Section~\ref{sec:related}, this makes graph-based indexing of a KV cache attractive only in settings where the same KV cache is reused multiple times.
In contrast, for \OURS, pre-filling the index takes a few seconds on modern GPUs.

\subsection{Ablation experiments}%
\label{sec:ablations}

We have evaluated numerous more complex variants that did not yield consistent improvements: 
increasing the model capacity, 
training a key classification model, 
fine-tuning the query classification model to the current prompt, 
using causal masking in the training batches, 
training on book summarization data rather than books to force long-range queries. 
In general, we observed that good results from experiments at the level of one head do not guarantee an improved end-to-end performance. 

\mypar{Q-model architecture and training.}
In Appendix~\ref{app:archablation}, we analyze some variants of the Q-model training. 
This analysis shows that using long-range queries-key pairs for training is useful. 
Using a residual architecture rather than a simple MLP does not help. 

\mypar{Dense context size.}
\OURS's default dense context is 2048 tokens (the attention sink of size 1 plus the 2047 most recent keys). 
We experiment with some other settings for attention sink+dense context: 1+511, 1+63 and 128+512. 
The results in Appendix~\ref{app:densecontext} show that the performance remains similar. 

\mypar{Batched queries.}
\OURS combines all four queries belonging to the same group into a single search, see Section~\ref{sec:implem}. 
These queries could also be performed independently. In Appendix~\ref{app:batchedqueries}, we show that batching the queries corresponding to the same KV head outperforms ranking the buckets independently for each query.

\mypar{Number of clusters $C$ and probes $\nprobe$.} We discuss the trade-off in Appendix~\ref{app:numberofclusters}. The selectivity is roughly proportional to $\nprobe/C$. Large $C$ provides better quality but is limited for efficiency reasons. 

\subsection{Limitations}
\label{sec:limitations}

Our method involves an additional training stage. While its computational cost is negligible compared to the LLM training, it involves some expertise (1) to ensure that the model quality is not significantly impacted for the target use-case(s), (2) to reset the hyper-parameters if the ones we report are not adapted for the target model. 
Another limitation is that our method is a solution for inference, not to training and not for prefill (although we do not see any major obstacle to adopt it for prefill).

\section{Conclusion}

We introduced \OURS, a new method for non-exhaustive attention computations in LLMs. 
\OURS is data adaptive, while incurring negligible run-time or memory overhead compared to exhaustive attention. 
Our paper has focused on the generation stage, where memory I/O has the strongest impact on performance. 
We have shown the efficiency of \OURS for typical long-context tasks, with significant computation speedups and a small impact on accuracy. 

\pagebreak

\bibliographystyle{assets/plainnat}
\bibliography{references}

\appendix

\begin{center}

\scalebox{0.95}{\hspace{-0.5em} \LARGE \textbf{Appendices}} \\[0.2em]
\medskip
\end{center}

In the following appendices, we present a few additional results that complement the main paper. 
Appendix~\ref{app:trainingdetails} provides details regarding the training procedure. 
Appendix~\ref{app:lshvariants} examines a few LSH-based partitioning methods. 
Appendices~\ref{app:rulerresults} and \ref{app:infinitebenchresults} report experimental results for RULER and additional ones for Infinitybench. 
Appendix~\ref{app:ablations} reports the results of ablation experiments. 
Appendix~\ref{app:algorithm_details} details how the \OURS algorithm is implemented on GPU. 

\section{Training details}
\label{app:trainingdetails}

The training was performed on 40 prompts of 500k tokens each. 
For these we have access to all key and query vectors from all layers and heads.

\mypar{Key assignment.} we train the k-means for 10 iterations with a random initialization. 
This is standard setting for partition-based indexing methods, we use the IVF implementation of the Faiss library~\citep{douze2024faiss}. 

\mypar{Q-model.} We train the Q-model for 40k iterations. 
At each iteration, we sample 1000 queries vectors from one random prompt. 
The query vectors are sampled among queries for which the nearest key in the prompt is more than 1024 tokens away in the sequence, to favor long-range queries (around 21\% the queries have this property). 
All the key vectors from the same prompt are used for training, except the attention sink. 

The loss of Equation~\ref{eq:kldiv} is optimized with the Adam optimizer and a learning rate of $10^{-5}$.
Note that we set a high number of iterations to get the best possible setting and because the model's training time is not critical for the pre-fill or inference speed.

\section{Partitioning the KV-cache with LSH}
\label{app:lshvariants}

\newcommand{\igN}[1]{\includegraphics[width=0.25\linewidth,trim={1cm 0 0 0},clip]{figs/LSH_comparison/LSH_variants_#1.pdf}}
\begin{figure*}[b]
    \centering
    \includegraphics[height=0.25\linewidth,trim={0 0 0 0},clip]{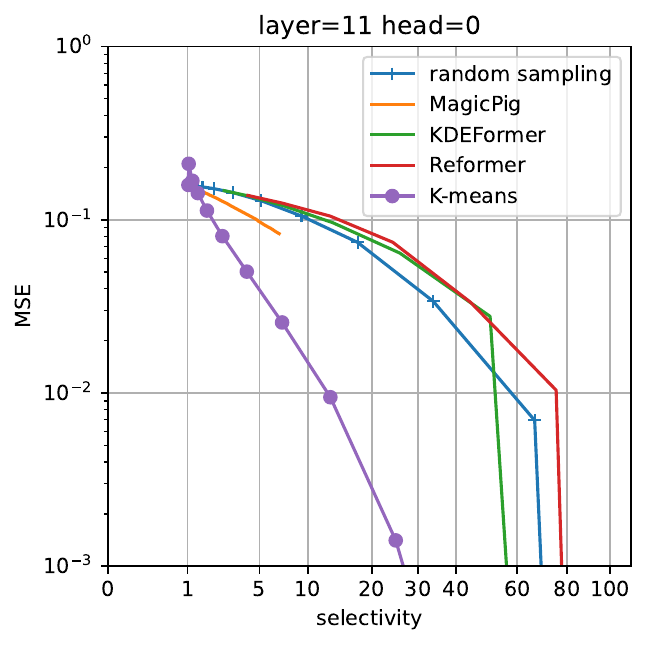}%
    \hfill
    \includegraphics[height=0.25\linewidth,trim={0.6cm 0 0 0},clip]{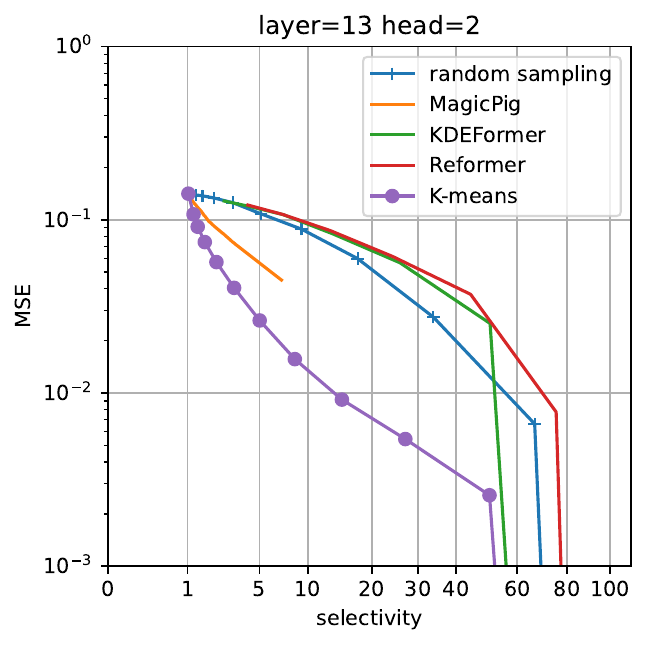}%
    \hfill
    \includegraphics[height=0.25\linewidth,trim={0.6cm 0 0 0},clip]{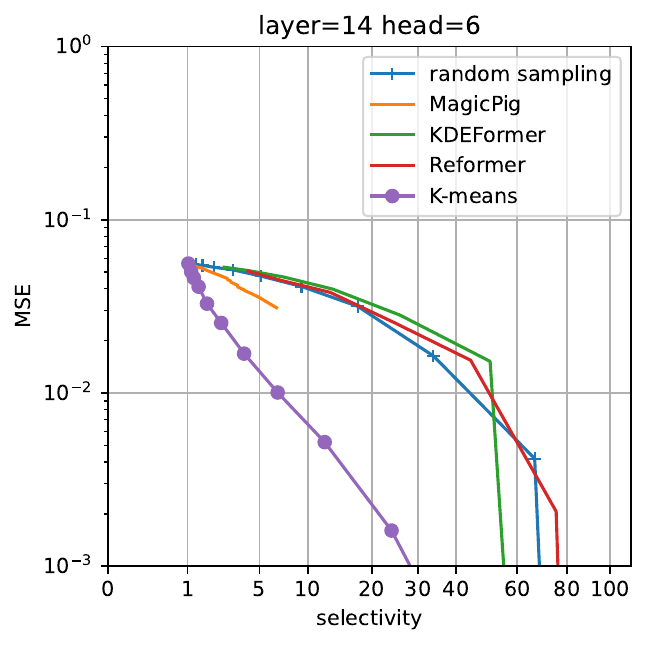}%
    \hfill
    \includegraphics[height=0.25\linewidth,trim={0.6cm 0 0 0},clip]{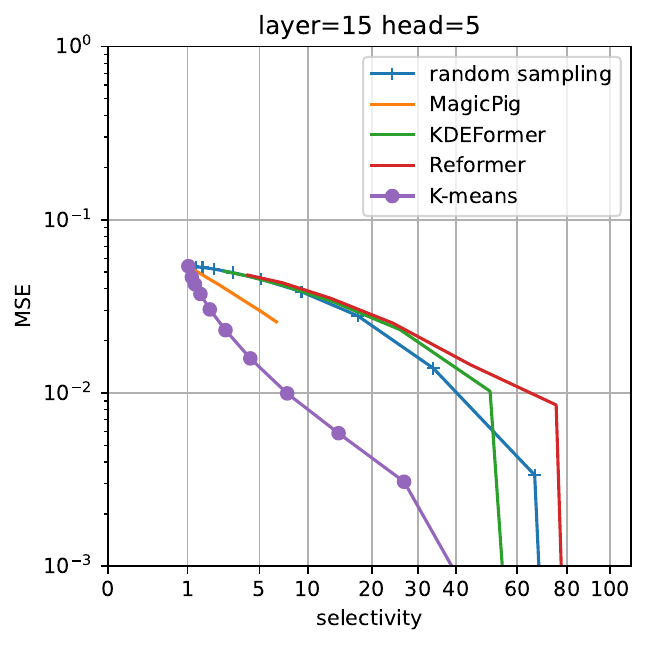}   
    \caption{
MSE vs. selectivity for three LSH variants, compared to a random selection and a partition based on k-means (the baseline for \OURS{}), on 4 representative heads. 
    }
    \label{fig:LSHvariants}
\end{figure*}

Locality Sensitive Hashing refers to multiple algorithms inspired by the Johnson-Lindenstrauss lemma. In the context of approximate nearest neighbor research, the term is overloaded, as it covers several algorithms and variants.
One common component is that the vector to hash $x\in\mathbb{R}^d$ is projected onto $r$ directions drawn randomly, which yields the $r$-bit bucket number:
\begin{equation}
    B(x) = \sum_{i=0}^{r}
    2^i \mathbf{1}[x^\top\pi_i > 0] \in \{0,\dots,2^r - 1\}. 
    \label{eq:lshone}
\end{equation}

\paragraph{Multiple hash tables.}

MagicPig~\citep{zhuoming2024magicpig} employs \emph{several} (up to $L=200$) hash tables. The vector search relies on collisions between the buckets: a key is considered only if it is hashed together with the query in at least two hash tables. 
Using such a large number of hash tables is classical for LSH, since a single hash table is very unbalanced. 
The theoretical guarantees of LSH apply only when the number of hash tables is large. 

Therefore, it is not strictly speaking a partitioning method of the keys. 
Besides, the large number of hash tables means that the memory used for that indexing structure is larger than the KV-cache itself ($4L = 800$ bytes per vector to store, compared with the $2\times2\times 128=512$ bytes to store the key-value pairs). 
This is why MagicPig stores the hash tables in CPU RAM. 

\paragraph{A single hash table?}

In the KDEformer method of~\citet{zandieh2023kdeformer},  a single hash table is considered, with a large $r$, so that the buckets are sparse. 
The bucket numbers corresponding to they keys $B(k)$ are ordered by their Gray codes\footnote{\url{https://en.wikipedia.org/wiki/Gray_code}} (which the authors re-invent).
Then the sequence of keys ordered in this way is partitioned in equal sized bins.  
At query time, only the bin the query vector ``falls'' in is visited. 

Since successive binary strings ordered by Gray codes differ by only 1 bit, the hope is that nearby buckets end up being nearby in this ordering.  
However, it is not guaranteed that if $B(k)$ and $B(k')$ differ by only 1 bit, then they will be nearby in this enumeration. 
Indeed, the bit that they differ by is not necessarily the one that changes in successive Gray codes. 
Besides, it is \emph{also} unlikely that many keys differ by only one bit because if $r$ is large then most vectors will differ by more than one.

\paragraph{Alternative hash functions.}

In the Reformer of~\citet{kitaev2020reformer}, the hashing function is replaced with the maximum dot product of random directions:
\begin{equation}
    B(x) = \underset{i=1..2^r}{\mathrm{argmax}}~~ x^\top\pi_i. 
\end{equation}
Compared to Equation~(\ref{eq:lshone}), each random projection only generates a single bucket instead of a bit for the bucket number.  
This is close to our k-means approach, if the centroids were drawn randomly. 

Similar to KDEFormer, the sequence of keys is sorted and split arbitrarily, but the bucket numbering does not have a significance here. 
The authors observe that it is beneficial to use multiple hash tables. 

\paragraph{Discussion.}

The theoretical grounding of these LSH based methods is fragile. 
They rely on a supposed property explicited in the HyperAttention work~\citep{han2023hyperattention}:
\begin{quote}
{\em
A useful property of this LSH variant is that its buckets are ordered in such a way that geometrically
adjacent buckets have consecutive buckets.    }
\end{quote}
There is an error in this reasoning: it is not possible to map a high-dimensional space to a sequence while maintaining neighborhood relations; otherwise, nearest neighbor search in high dimensions would be as easy as in 1D. 

Besides, the subsequent arbitrary splitting of linear sequences into regular buckets defined to minimize computations defeats the purpose of the original data buckets. 

\paragraph{Experiments.}

We compare different LSH approaches in the tradeoff between selectivity and mean squared error (MSE) for a few attention heads. 
The MSE is computed between the approximate attention output and the exact one. 
We use random sampling as the baseline. 
Since the attention mechanism can be seen as Gaussian kernel smoothing~\cite{zhang2024memory}, random sampling is actually a reasonable estimator, that reaches MSE=0 when all vectors are sampled (we use sampling without replacement). 

All the variants have a parameter that adjusts the tradeoff:  
for KDEFormer and Reformer, fewer larger buckets improve the accuracy and increase the computational cost (1 bucket covering all keys and queries is equivalent to full attention); 
for MagicPig's LSH with multiple hash tables, increasing the number $L$ of hash tables improves the coverage of the key vectors. 

Figure~\ref{fig:LSHvariants}  shows that for this metric, the KDEFormer and Reformer are on-par or slightly worse than random sampling. 
Note, however, that these two paritioning methods were developed to be used at pre-fill time, so their performance could be better in a self-attention setting where the contexts embeddings are matched among themselves. 
In fact, Reformer uses the query corresponding to a key to assign that key to a bucket instead of attempting to assign the key itself. 
MagicPig implements the classical multi-hashtable LSH, so it performs better, but note that since MagicPig uses multiple hash tables, it does not yield a (single) partition of the dataset. 
The k-means approach obtains much better results, mainly because it is data-adaptive. 

These LSH based methods are confronted with the same bucketing balancing issues that we encounter in \OURS.
MagicPig resolves this balancing issue by running the search on CPU, that is more tolerant to irregular computation patterns than GPUs. 
The two other variants arbitrarily split the keys and queries into fixed-size buckets that a handled easily by block computations.  
\OURS attempts to balance the buckets at training time with a specific loss term.

\section{RULER results}
\label{app:rulerresults}

In~\autoref{tab:RULER}, we report the results of the RULER~\citep{hsieh2024rulerwhatsrealcontext} subtasks: multi-key, multi-query and VT for two context lengths 65k and 128k. 
Surprisingly, the SAAP scores are often above the full attention scores. This can be explained by the fact that the sparse attention focuses on only a relevant subset of the prompts, thus it has a denoising effect. 
SAAP scores are also higher than k-means in three out of five tasks.

\begin{table}[h]
    \centering
    \caption{
    Scores with a Llama 3.1-8B model using rope extrapolation to increase the context length of 500k and different partitioning schemes. We report in \textbf{bold} the best results among approximate attention schemes. Each subtask  has the corresponding context size  in parentheses.
    }
    \smallskip
\scalebox{0.83}{
\noindent 
  \begin{tabular}{@{\ }l|rrrrr@{\ }}
    \toprule
    &\multicolumn{5}{c}{RULER~\citep{hsieh2024rulerwhatsrealcontext}} \\
 Methods& \multicolumn{2}{c}{multi-key}	&	\multicolumn{2}{c}{multi-query} &	VT \\
context length & 65k & 128k & 65k & 128k & 65k \\
    \midrule
\textcolor{gray}{Full attention}  & \textcolor{gray}{39.0}&	\textcolor{gray}{25.8}	&\textcolor{gray}{54.5}	&\textcolor{gray}{62.7}	&\textcolor{gray}{50.5 }\\
StreamingLLM (1+2047) & 3.6	&2.0	&1.9	&1.7	&5.0  \\
K-means (roped inputs, 
$\nprobe=32$) & 27.6	&14.2	&\textbf{78.4}	&60.7&	78.4\\
K-means (no rope,
$\nprobe=32$) & 34.4	&24.2	&65.0	&\textbf{67.7}&	54.2 \\
\OURS, K-means+Q ($\nprobe=32$)
& \textbf{46.6}	&\textbf{31.8}&	70.0&	65.7	&\textbf{85.0}\\
\bottomrule
  \end{tabular}\label{tab:RULER}}
   \bigskip  
\end{table}

\section{InfiniteBench results}
\label{app:infinitebenchresults}

\begin{figure*}
    \includegraphics[height=0.25\linewidth]{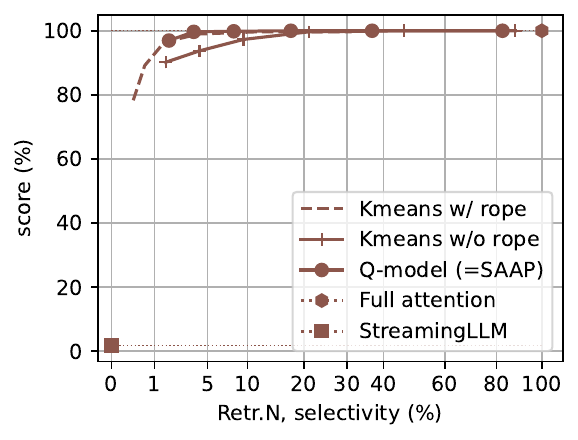}%
    \hfill
    \includegraphics[height=0.25\linewidth,trim={0.6cm 0 0 0},clip]{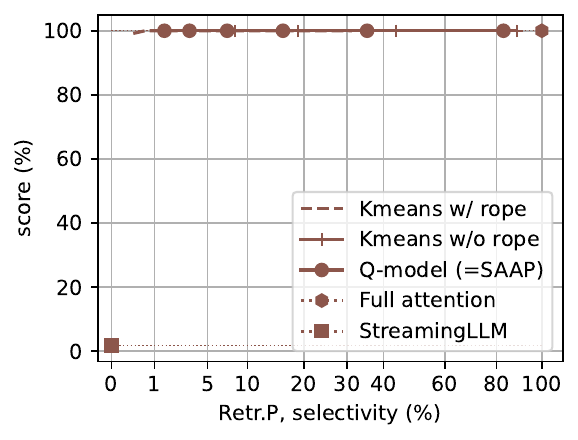}
    \hfill
    \includegraphics[height=0.25\linewidth,trim={0.6cm 0 0 0},clip]{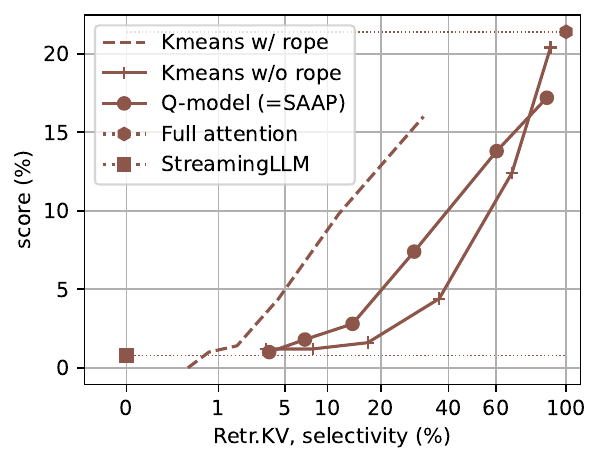} \\%
    \includegraphics[height=0.25\linewidth]{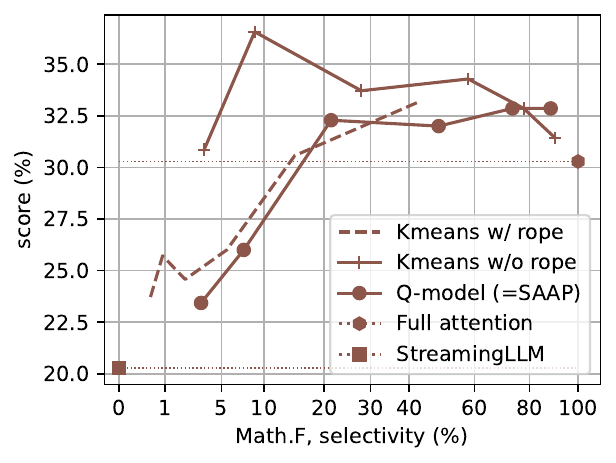}
    \hfill
    \includegraphics[height=0.25\linewidth,trim={0.6cm 0 0 0},clip]{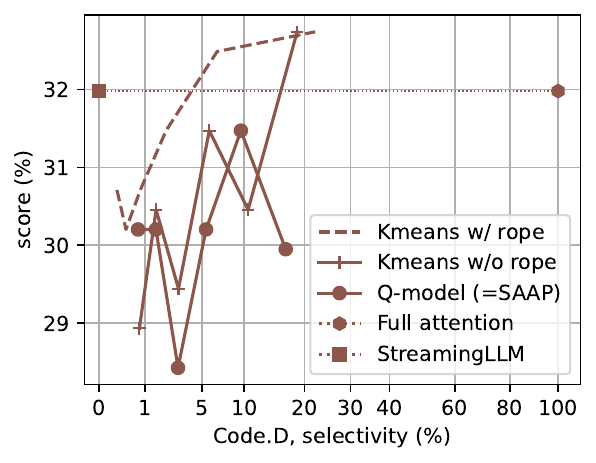}%
    \hfill
    \includegraphics[height=0.25\linewidth,trim={0.6cm 0 0 0},clip]{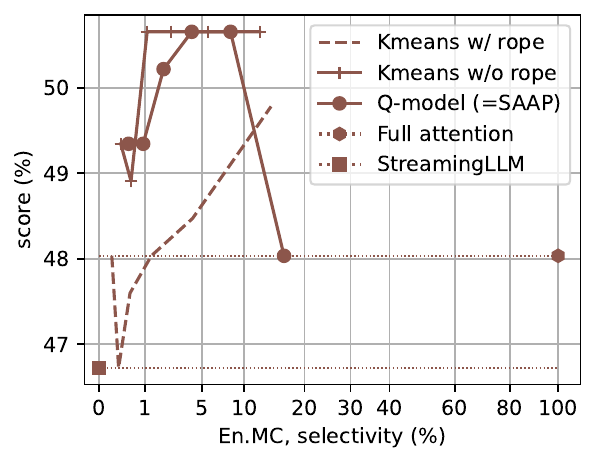}    
    \caption{
        Performance on selected InfiniteBench tasks 
    \label{fig:ibres}}
\end{figure*}

\paragraph{Additional results.} \autoref{fig:ibres} complements ~\autoref{fig:accuracyatbudget} and illustrates the task-dependent behaviors described in Section~\ref{sec:baselines}.
\label{app:infinitebenchres}. 
The results are more noisy, yet \OURS does perform well for most tasks. 

\section{Ablation experiments}
\label{app:ablations}

In these experiments we evaluate our method with variations of a few key parameters. 

\subsection{Variations on architecture and training setting}
\label{app:archablation}

\begin{figure}
    \includegraphics[height=0.39\linewidth]{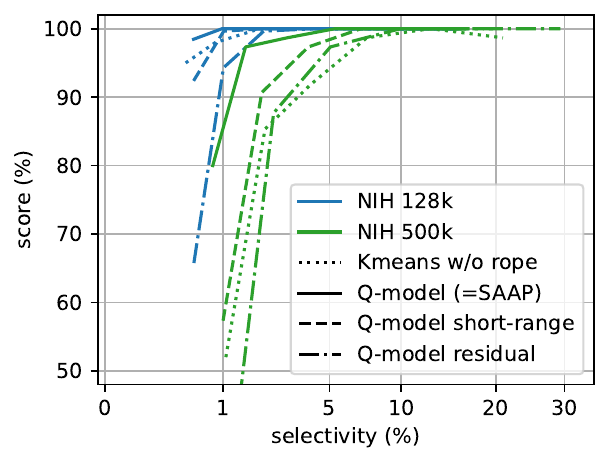}%
    \hfill
    \includegraphics[height=0.39\linewidth,trim={0.7cm 0 0 0},clip]{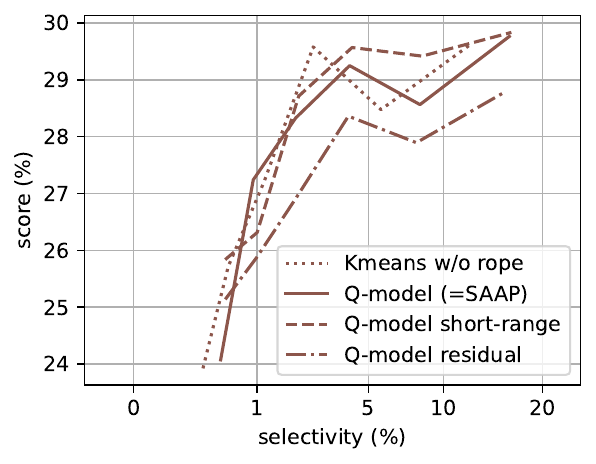}  
    
    \caption{
        Comparison of architectural variants on the needle-in-haystack (left) and InfiniteBench En.QA (right) tasks. 
    }
    \label{fig:nih-arch}
\end{figure}

Figure~\ref{fig:nih-arch} shows the performance on the NIH and En.QA tasks using different architectures.
It is comparable to Figure~\ref{fig:accuracyatbudget} with a few more variants for the Q model: 
\begin{itemize}
    \item 
    ``Q-model short-range'' is the standard 2-layer MLP, but trained without filtering out short-range queries ;
    \item 
    ``Q-model residual'' relies on a residual architecture with the same capacity as the MLP, also trained without filtering out short-range queries.
\end{itemize}

The NIH plot shows that training on long-range queries only improves the accuracy significantly. 
Besides, the residual architecture is not an improvement over the MLP. 
The En.QA results are more contrasted but the task is more noisy, so we chose to standardize on the MLP architecture with short-range filtering. 

\subsection{Dense context size}
\label{app:densecontext}

\begin{figure}
    \includegraphics[height=0.39\linewidth,trim={0.0cm 0 0 0},clip]{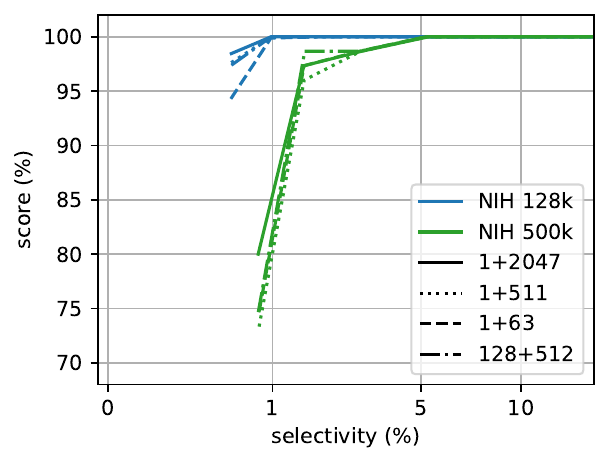}%
    \hfill    
    \includegraphics[height=0.39\linewidth,trim={0.7cm 0 0 0},clip]{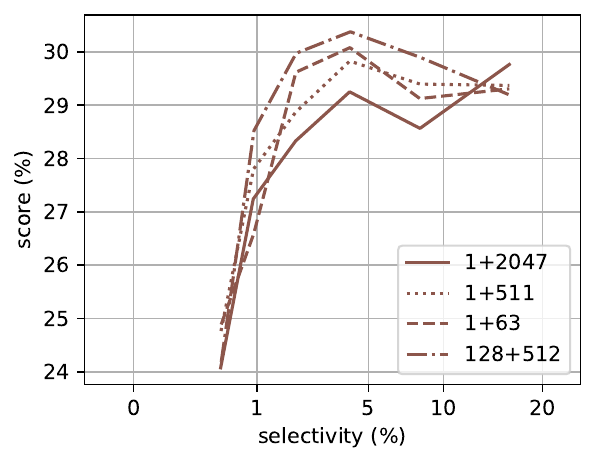}   \\    
    \caption{
        Comparison of dense context sizes on the needle-in-haystack (left) and InfiniteBench En.QA (right) tasks. 
        The dense context is reported as ``128+512'' means that the first 128 tokens of the sequence and the most recent 512 ones are accessed exactly (dense). 
    \label{fig:nih-dense}}
\end{figure}

Figure~\ref{fig:nih-dense} shows the performance of various dense context sizes.
Depending on the task, some dense context settings are more efficient. 
We chose 1+2047 that works well on the needle-in-haystack task. 

\subsection{Batched queries}
\label{app:batchedqueries}

\begin{figure}
    \includegraphics[height=0.38\linewidth]{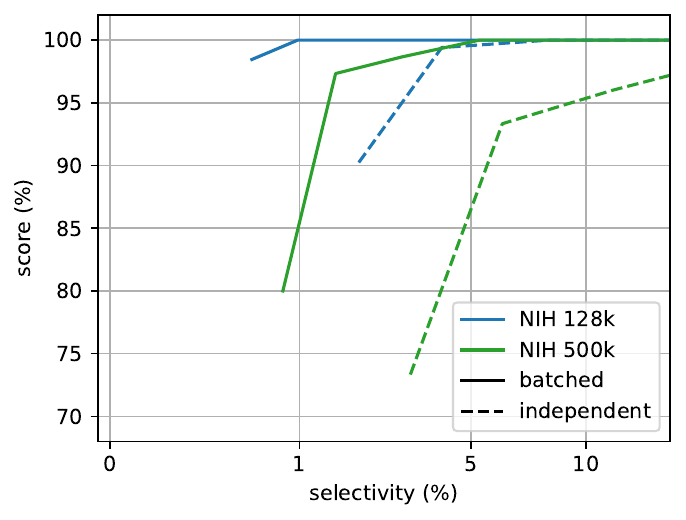}%
    \hfill
    \includegraphics[height=0.38\linewidth,trim={0.7cm 0 0 0},clip]{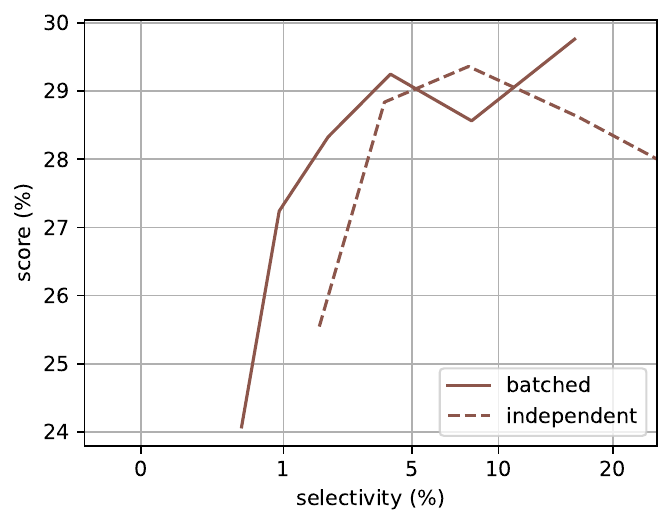}  
    \caption{
        Comparison of batched and independent queries on the needle-in-haystack (left) and InfiniteBench En.QA (right) tasks. 
    }
    \label{fig:nih-batched}
\end{figure}

Figure~\ref{fig:nih-batched} shows the effect of batching the 4 queries belonging to the same KV head.

\begin{itemize}
    \item 
    "Batched": Jointly rank partitions across multiple queries to identify the top-k partitions that we want to access.
    \item 
    "Independent": Access the top-k ranked partitions for each query separately.
\end{itemize}
Interestingly, we observe that the batched setting often outperforms the independent setting, which ought to be more data-adaptive. 

\subsection{Guidance for the number of clusters and the number of visited clusters}
\label{app:numberofclusters}

We carried out additional experiments with a varying number $\nlist$ of partitions in a setting similar to Figure~\ref{fig:LSHvariants}: we measure the MSE (approximation error) with k-means partitioning (for a few arbitrary layers and heads). 
We vary the number of visited clusters 
$\nprobe$ for several settings of $\nlist$. 
The selectivity roughly is proportional to $\nprobe/\nlist$.

\begin{figure}
    \includegraphics[height=0.39\linewidth]{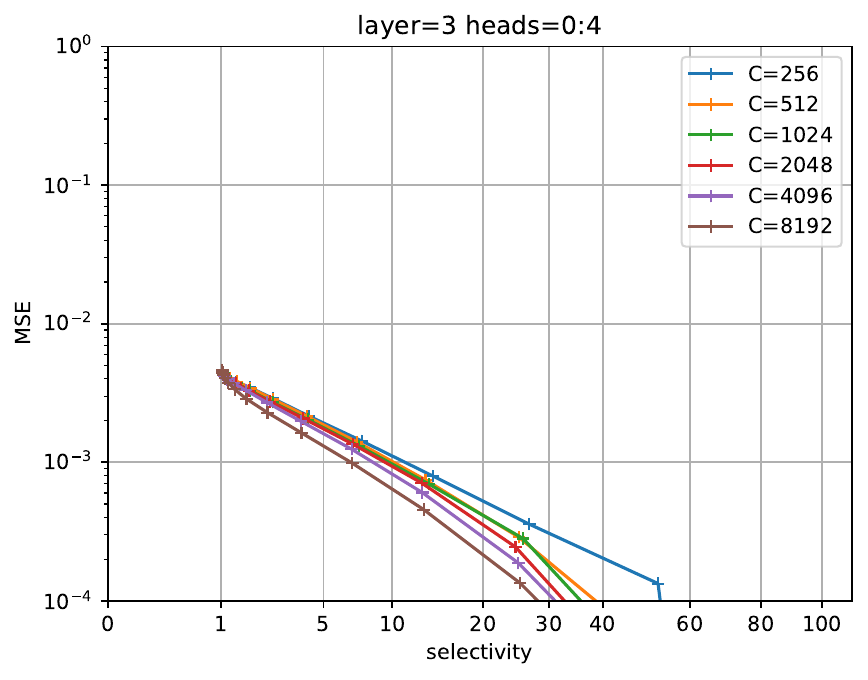} 
\hfill
    \includegraphics[height=0.39\linewidth,trim={0.7cm 0 0 0},clip]{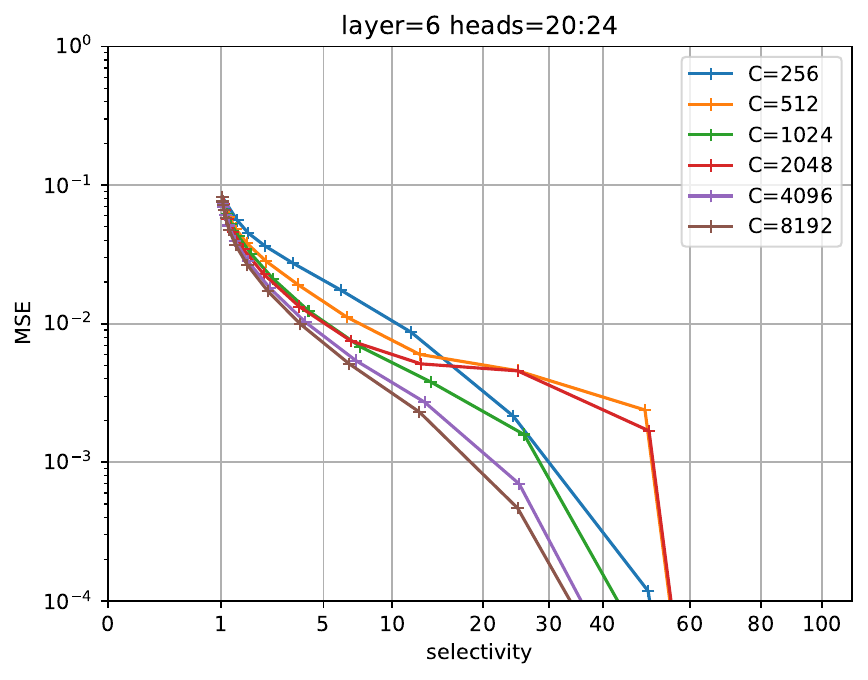} 
\\
    \includegraphics[height=0.39\linewidth]{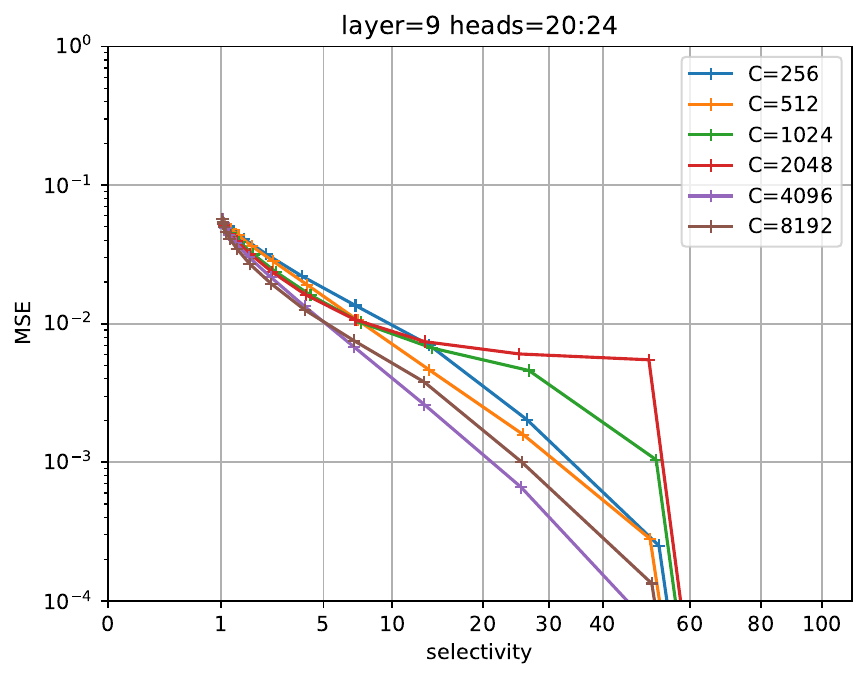} 
\hfill
    \includegraphics[height=0.39\linewidth,trim={0.7cm 0 0 0},clip]{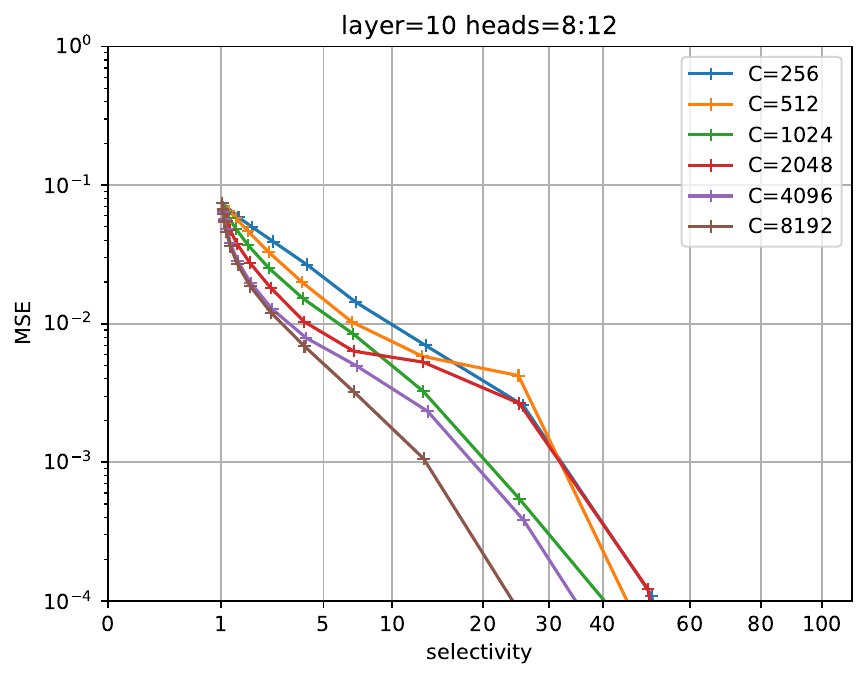} 
    \caption{Error (MSE) as a function of the number of clusters $\nlist$ and the number of probes $\nprobe$.}
    \label{fig:MSEvsSel}
\end{figure}

Figure~\ref{fig:MSEvsSel} shows that with a fixed $\nprobe/\nlist$ and selectivity, a lower (better) MSE can be obtained with a larger number of clusters $\nlist$. 
This is also a classical observation in partition-based indexing~\citep{douze2024faiss}.
However, a larger $\nlist$ poses several problems: 
(1) the assignment operation is more expensive, 
(2) it is harder to train – especially for a Q-model, 
(3) the GPU implementation becomes less efficient because the memory regions to access are more fragmented. 
Therefore, we set $\nlist$=1024 by default. 

\begin{figure}
    \centering
    \includegraphics[height=0.6\linewidth]{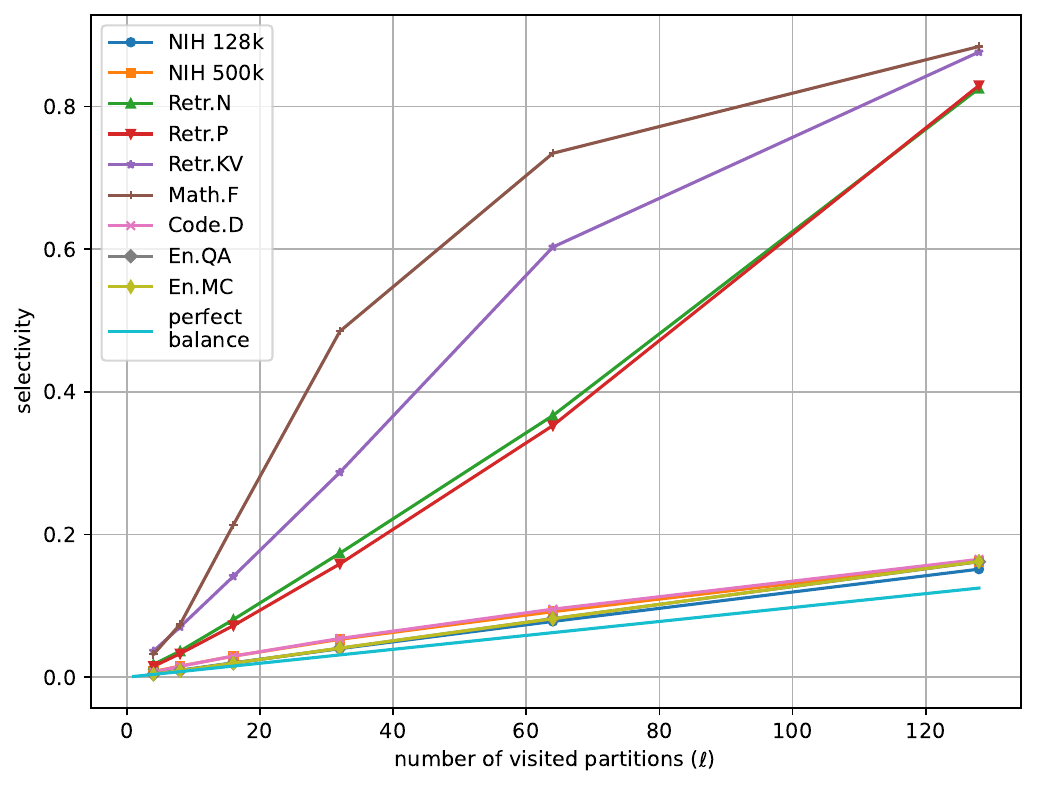}
    \caption{Selectivity at various choices for $\nprobe$, the number of clusters visited.
    }
    \label{fig:nprobe_selectivity}
\end{figure}

Selectivity is a monotonously increasing function of the hyperparameter $\nprobe$, the number of clusters visited per query. 
\autoref{tab:nprobe_selectivity} and \autoref{fig:nprobe_selectivity} show that depending on the task, the selectivity increases faster with $\nprobe$ , eg. for NIH the selectivity at $\nprobe = 32$ is around 4\% while it is 48.5\% for Math.F. This high selectivity can be explained by the fact that the Math.F benchmark is built on long sequences of numbers, which is a different domain compared to natural language. For this task \OURS is ineffective, because at this selectivity level, the speedup with respect to full attention is not significant.

By contrast, for Code.D the selectivity is close to that of natural language task en.QA. This is interesting because, while the LLM training data includes code, the key and query partitioning models did not: in this case \OURS is relatively robust. 

The relationship between $\nprobe$ and selectivity can also be observed in the curves of \autoref{fig:accuracyatbudget}, where the score vs. selectivity operating points are obtained by varying $\nprobe = 4, 8, ..., 256$.

\newcommand{\igM}[1]{\includegraphics[width=0.32\linewidth]{figs/MSE_vs_selectivity/MSE_vs_selectivity_#1.pdf}}

\begin{table}
    \caption{
        Selectivity when varying the number $\nprobe$ of visited clusters. 
        \label{tab:nprobe_selectivity}
    }
{\centering 
\scalebox{0.8}{
\begin{tabular}{@{\ }l|rrrrrrrrrr@{\ }}
\toprule
$\ell$ & \rotatebox{75}{balanced} & \rotatebox{75}{NIH 128k} & \rotatebox{75}{NIH 500k} & \rotatebox{75}{Retr.N} & \rotatebox{75}{Retr.P} & \rotatebox{75}{Retr.KV} & \rotatebox{75}{Math.F} & \rotatebox{75}{Code.D} & \rotatebox{75}{En.QA} & \rotatebox{75}{En.MC} \\
\midrule
4 & 0.4 & 0.5 & 0.8 & 1.8 & 1.5 & 3.7 & 3.2 & 0.7 & 0.4 & 0.4 \\
8 & 0.8 & 1.0 & 1.5 & 3.7 & 3.3 & 7.0 & 7.4 & 1.5 & 0.9 & 0.9 \\
16 & 1.6 & 2.0 & 3.0 & 8.1 & 7.2 & 14.1 & 21.4 & 3.0 & 2.0 & 2.0 \\
32 & 3.1 & 4.0 & 5.3 & 17.4 & 15.9 & 28.7 & 48.5 & 5.4 & 4.1 & 4.1 \\
64 & 6.2 & 7.8 & 9.2 & 36.7 & 35.3 & 60.3 & 73.5 & 9.5 & 8.2 & 8.2 \\
128 & 12.5 & 15.1 & 16.2 & 82.6 & 83.0 & 87.7 & 88.4 & 16.5 & 16.2 & 16.2 \\
\bottomrule
\end{tabular}}}
\end{table}

\section{Algorithm details}
\label{app:algorithm_details}

In this section, we will describe the algorithm details of the \OURS Triton kernel.
Given a query $Q \in \mathbb{R}^{1 \times d}$, 
the 1+2047 keys and values that are accessed exactly are $K_l, V_l \in \mathbb{R}^{N_l \times d}$;
the remaining keys and values are $K_s, V_s \in \mathbb{R}^{N_s \times d}$, %
Every $K_s, V_s$ has a corresponding partition assignment $A_i \in \{1, ... C \}$ for $i \in \{ 1, ..., N_s \}$, where $C$ is the number of buckets.

\newcommand{\Off}{\texttt{Off}}
\newcommand{\Idx}{\texttt{Idx}}
\newcommand{\PartialAttention}{\texttt{PAttn}}
\newcommand{\rowmax}{\text{rowmax}}

From this partition assignment, we build an index comprised of two tensors $\Off, \Idx$ that groups all assignments to the same bucket together, and allows for efficiently iterating over all elements in a bucket. 
This representation is similar in principle to a compressed sparse-row representation, where $\Off$ (of size $C$) would map to the compressed rows and $\Idx$ (of size $N$) to the column indices.
Thus, the list of key/value ids within bucket $c$ is $\Idx[\Off[c]:\Off[c+1]]$, where $i:j$ denotes indices $\{i,\dots,j-1\}$.

The key and value vectors are stored in HBM (high-bandwidth memory) that is \emph{slower} than the on-chip SRAM.

For a given query $Q$, we assign the $\nprobe$ most promising buckets in $B \in \{1, ..., C\}^{\nprobe}$ .

We parallelize the program by launching $\nprobe + 1$ CUDA blocks, where the first block will handle the 1+2047 $K_l, V_l$, and each of the remaining blocks will handle a given bucket and will perform the sparse attention. Our kernel assumes the grouped-query attention setting where keys and values share the same head, with only the queries having $n_h$ independent heads. Given this setting, we avoid duplicating the $K, V$ heads and we compute the $n_h$ heads from $Q$ together, by leveraging the fact that the sequence length for $Q$ is 1, so that the computation is equivalent to transposing the head and the sequence dimension for $Q$.

\begin{algorithm}
\caption{Partial attention kernel \PartialAttention}
\label{algo:partial_attention}
\begin{algorithmic}%
\small
\REQUIRE Inputs $Q, O^* \in \mathbb{R}^{n_h \times d}$, $K, V \in \mathbb{R}^{B_c \times d}$, $m^* \in \mathbb{R}^{n_h \times 1}$ and $s^* \in \mathbb{R}^{n_h \times 1}$ all in on-chip SRAM
\STATE Compute $s := Q K/\sqrt{d}$
\STATE Compute $m_i := \rowmax (m^*, A)$
\STATE Update $O^* := O^* \cdot \exp(m^* - m_i) + V \exp(s - m_i)$
\STATE Update $s^* := s^* \cdot \exp(m^* - m_i) + \exp(s - m_i)$
\STATE Update $m^* := m_i$
\end{algorithmic}
\end{algorithm}

\begin{algorithm}
\caption{Inverted list computation kernel}
    \begin{algorithmic}%
    \small
\REQUIRE Inputs $Q \in \mathbb{R}^{n_h \times d}$, $K_s, V_s \in \mathbb{R}^{N_s \times d}$, $K_l, V_l \in \mathbb{R}^{N_l \times d}$, \\
buckets to visit: $B \in \{1, ..., C\}^\ell$, \\
$\Off$ and $\Idx$, \\
key block size $B_c$, and CUDA block id $b_i$.\\
\STATE Load $Q$ from HBM to on-chip SRAM
\STATE Initialize accumulator $O_{b_i} := 0_d$
\STATE Initialize sumexp accumulator $s_{b_i} := 0$
\STATE Initialize max accumulator $m_{b_i} := -\infty$
\IF {$\ell >= b_i > 0$}
\STATE Read $c=B[b_i - 1]$, the bucket id to visit
\STATE Let $I^c$ be the indices of $\Idx$ that correspond to bucket $c_i$ by indexing $I^c = \Idx[\Off[c]:\Off[c+1]]$
\STATE Divide the indices $I^c$ in $T = \lceil \frac{|I^c|}{B_c} \rceil$ blocks $\{I^c_1, ... I^c_{T}\}$ of size $B_c$
\FOR{$0 \le j < T$}
\STATE Load indices $I_j^c$ to SRAM.
\STATE Strided load $K_s^j, V_s^j := K_s[I_j^c], V_s[I_j^c]$ to SRAM
\STATE $\PartialAttention(Q, K_s^j, V_s^j, m^{b_i}, s_{b_i}, O_{b_i})$ %

\ENDFOR
\ELSIF {$b_i == 0$}
\STATE compute block size $T = \lceil \frac{N_l}{B_c} \rceil$
\FOR{$0 \le j < T$}
\STATE compute indices $i_0=B_cj$ and $i_1=B_c(j+1)$
\STATE  $\PartialAttention(Q, K_l[i_0:i_1], V_l[i_0:i_1], m^{b_i}, s_{b_i}, O_{b_i})$ %

\ENDFOR
\ENDIF
\STATE Write $O_{b_i}, s_{b_i}, m_{b_i}$ to HBM
\\
\hrulefill
\\
\STATE {\bf In a separate kernel, for every $n_h$ compute merge-attention}:

\STATE Load $O_{0}, s_{0}, m_{0}$
\STATE Set $m^* := m_{0}$, $O = O_{0}$ and $s = s_{0}$
\FOR{$1 \le i \le \nprobe$}
\STATE Load $O_{i}, s_{i}, m_{i}$
\STATE Compute $\hat{m} := \max(m^*, m_{i})$
\IF {$m_{b_i} < m^*$}
\STATE Compute $\alpha := \exp(m_{i} - \hat{m})$
\STATE Update $O_{i} := \alpha O_{i}$
\STATE Update $s_{i} := \alpha s_{i}$
\ELSE
\STATE Compute $\alpha := \exp(m^* - \hat{m})$
\STATE Update $O := \alpha O$
\STATE Update $s := \alpha s$
\ENDIF
\STATE Set $m^* := \hat{m}$, $s := s + s_{i}$ and $O := O + O_{i}$
\ENDFOR
\RETURN $O / s$

    \end{algorithmic}
\end{algorithm}

Once the partial results are computed, we apply the merge-attention function from xFormers~\citep{xFormers2022} to aggregate the partial results into the final output.

\end{document}